\documentclass[11pt, a4paper, logo, onecolumn, copyright]{deepmind}

\usepackage{caption}
\DeclareCaptionLabelSeparator{pipe}{\textbar}
\usepackage[utf8]{inputenc}
\usepackage{mdframed}
\usepackage{xcolor}

\usepackage{graphicx}
\graphicspath{ {./figures/} }

\title{A social path to human-like artificial intelligence}

\author[1]{Edgar~A.~Du\'e\~nez-Guzm\'an}
\author[2]{Suzanne~Sadedin}
\author[1]{Jane~X.~Wang}
\author[1]{Kevin~R.~McKee}
\author[1]{Joel Z.~Leibo}

\affil[1]{Google DeepMind}
\affil[2]{Independent researcher}

\correspondingauthor{Edgar~A.~Du\'e\~nez-Guzm\'an: duenez@google.com}

\keywords{artificial intelligence, multi-agent systems, social intelligence, cultural intelligence, major transitions.}

\begin{abstract}
    \noindent Traditionally, cognitive and computer scientists have viewed intelligence solipsistically, as a property of unitary agents devoid of social context. Given the success of contemporary learning algorithms, we argue that the bottleneck in artificial intelligence (AI) progress is shifting from data assimilation to novel data generation. We bring together evidence showing that natural intelligence emerges at multiple scales in networks of interacting agents via collective living, social relationships and major evolutionary transitions, which contribute to novel data generation through mechanisms such as population pressures, arms races, Machiavellian selection, social learning and cumulative culture. Many breakthroughs in AI exploit some of these processes, from multi-agent structures enabling algorithms to master complex games like Capture-The-Flag and StarCraft II, to strategic communication in Diplomacy and the shaping of AI data streams by other AIs. Moving beyond a solipsistic view of agency to integrate these mechanisms suggests a path to human-like compounding innovation through ongoing novel data generation.
\end{abstract}

\begin{document}

\maketitle

\begin{figure*}
  \centering
  \includegraphics[scale=0.9]{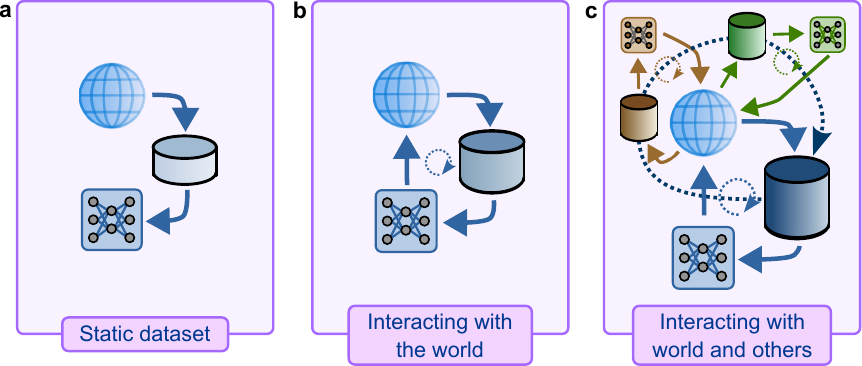}
  \caption{\textbf{Learning quality depends on the richness and size of the dataset.} Learning agents improve as a function of the richness and size of their dataset. \textbf{a}, A single agent learns from a large but static dataset. \textbf{b}, Interaction with the world enables the agent to generate its own data, but can trap the agent in local minima, limiting the dataset richness. \textbf{c}, Other agents enrich the dataset by changing the world with which an agent interacts. \label{fig:schematic}}
\end{figure*}

\noindent A key lesson in artificial intelligence (AI) history is that the extent to which intelligent behaviour emerges in learning systems scales with the size of the dataset used for training~\cite{krizhevsky2012imagenet, deng2009imagenet, kaplan2020scaling, bommasani2021opportunities, hoffmann2022training, fei2022searching}. Current large models are trained on vast datasets and obtain human-level performance on a wide variety of tasks, especially in natural language, but increasingly in multi-modal domains~\cite{bommasani2021opportunities, alayrac2022flamingo}. Though large, these training datasets are exogenous to the models they train; consequently, their learning is dependent upon, and limited to, relationships in the data (Fig.~\ref{fig:schematic}a). There may be limits to what can be achieved by training on any such static dataset, even if it is very large. For example, an algorithm trained on all of human knowledge before the performance of the double-slit experiment which catalysed the development of quantum physics~\cite{young1804bakerian} would have no way of predicting the outcome in advance. To overcome this, algorithms need to generate their own data. Reinforcement learning (RL) algorithms do this~\cite{sutton2018reinforcement}. They learn by interacting with their environment and observing new states reached through their own behaviour, in effect generating continually-growing datasets for themselves (Fig.~\ref{fig:schematic}b). RL agents running in simulation can generate data indefinitely, so if their cognitive abilities scale with dataset size, you would expect dramatic results from this approach. However, this is not what typically happens. In small and stationary worlds, RL agents converge to behaviours that repeatedly execute the same actions in the same states, generating no new data. When this happens their learning stagnates~\cite{schaul2019ray, ortega2021shaking}. In natural language processing, a wave of research is seeing performance improvement from models generating their own training data in sophisticated ways~(e.g.~\cite{huang2022large}). However, naively training a model on data generated by itself tends to remove the tails of the distribution, degrading the data stream produced and thus limiting subsequent learning ~\cite{shumailov2023curse}. In this article, we argue that next-level intelligent systems may move beyond this constraint by continually generating novel data in self-organising multi-scale, multi-agent interactions.

RL algorithms strengthen rewarding actions even if they are rarely emitted, effectively amplifying the most profitable available behaviour~\cite{sutton2018reinforcement}. This works well if the initial behaviour distribution covers all relevant states. However, when the world is large enough that the initial behaviour distribution covers only a relatively tiny subspace, an intelligent agent must invent wholly new behaviours which they are extraordinarily unlikely to produce by accident. The agent's data stream may change through its actions, but the waiting time until a complex rewarding behaviour is emitted by chance may be so long that no learning gradient is available, thus such changes do not motivate further exploration. One way to improve exploration in RL algorithms is to have the environment adapt to the agent~\cite{wang2019paired}. Another is to make exploration an explicit part of the goal~\cite{Portelas2020automatic}. In intrinsic motivation models, the latter is achieved by incorporating models of curiosity~\cite{linke2020adapting, oudeyer2007intrinsic} that seek novel states~\cite{pathak2017curiosity} or via agents that generate their own intrinsic goals~\cite{colas2022autotelic}. However, this leads to a trade-off: computation must be balanced between seeking new goals (exploration), and improving behaviours on existing goals (exploitation). Substantial research effort has been devoted to balancing this trade-off~\cite{ladosz2022exploration, jiang2023general}, and some bounds are known~\cite{kearns2002near, osband2019deep}. Some approaches attempt to evade this trade-off, for example by focusing on the intrinsic goals where they make most progress. But a pernicious difficulty remains: even if the designer can avoid the exploration-exploitation trade-off, they must still decide the direction of exploration.

\begin{figure*}
  \centering
  \includegraphics[scale=0.8]{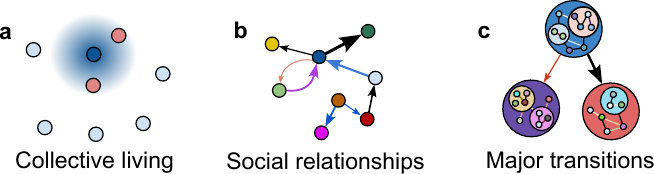}
  \caption{\textbf{Social interactions drive compounding innovation.} Interactions in the three drivers of compounding innovation have qualitatively different properties. \textbf{a}, Agent interactions in \emph{Collective living} are anonymous, mediated by proximity. \textbf{b}, In \emph{Social relationships}, the identity of individuals and their relationships matter during interactions, creating networks that facilitate cooperation and social learning. \textbf{c}, \emph{Major transitions} lead to the evolution of multi-scale agents, where larger-scale agents regulate the environments of smaller-scale ones.
  \label{fig:networks}}
\end{figure*}

By contrast, in the view we develop here, exploration and exploitation become synergistic rather than antagonistic. Exploitation drives exploration in the appropriate direction as innovation builds upon prior innovation. We call this effect \emph{compounding innovation}. Mechanistically, compounding innovation occurs when exploitation continuously generates new data that creates opportunities for learning: exploration \emph{by} exploitation~\cite{leibo2019autocurricula}. We argue compounding innovation arises in biological evolution because the environment continually changes with innovation due to specific types of interactions between biological units. These interactions gave rise to human cultural evolution---the process that generated the data on which our most intelligent algorithms have trained. Can we create analogous interactions in artificial learning systems? In a multi-agent system, the behaviour of any one agent influences the data streams of others (Fig.~\ref{fig:schematic}c), suggesting a possible path to ongoing learning. Yet contemporary multi-agent algorithms still fail at compounding innovation. Here, we consider three broad forms of social structure that arise in biological systems, each characterised by distinct types of social interaction (Fig.~\ref{fig:networks}) that may contribute to compounding innovation in different ways. First, we discuss the implications of ``collective living'': how competition between agents creates novel data by disrupting local equilibria, often generating naturally emergent sequences of learning opportunities that stimulate behavioural innovation~\cite{sukhbaatar2017intrinsic, leibo2019autocurricula, leibo2019malthusian, baker2019emergent, balduzzi2019open, openai2021asymmetric, goodfellow2014generative}. Second, we examine how social relationships facilitate cooperation between individuals and contribute to forms of cognition relevant to human-like behaviours, including social learning~\cite{herrmann2007humans, boyd2011cultural, whiten2019cultural} and Machiavellian selection~\cite{dunbar1998social, byrne2018machiavellian}. Finally, we discuss major evolutionary transitions~\cite{szathmary1995major, jablonka2014evolution} and their role in human cultural evolution via language~\cite{heyes2018cognitive}. Major transitions modulate interactions at multiple scales simultaneously, leading to units specialising, cooperating and competing as needed to solve goals at a higher level of abstraction. Learning and agency in such systems likewise emerge at multiple scales, allowing compounding innovation as low-level agency fine-tunes skills while higher-level agency recombines and coordinates them.

\section*{Collective living}

Group-living organisms commonly evolve social behaviour that enables decentralised groups to rapidly detect and respond to threats and opportunities, without requiring cooperation or individual relationships. Examples include bacterial quorum sensing~\cite{ng2009bacterial}, alarm signals in plants and animals~\cite{verheggen2010alarm}, and collective maze learning by rats~\cite{nagy2020synergistic}. As interactions between individuals become a source of emergent outcomes that change the adaptive landscape, collective living enables deeper innovation than would be possible in a single-agent system~\cite{schluter2000ecology}. In multi-agent systems, such interactions can provide autocurricula~\cite{leibo2019autocurricula}---naturally emerging data presentation sequences that may aid learning~\cite{bansal2017emergent, wang2019paired}.

Self-organisation allows individuals with only local information to exhibit adaptive behaviour at global scales. Examples in AI include swarm intelligence~\cite{reynolds1987flocks}, and multi-agent reinforcement learning systems modeling social dilemmas~\cite{lerer2017maintaining, leibo2017multiagent,mckee2022quantifying}, coordination~\cite{strouse2021collaborating}, communication~\cite{lazaridou2016multi}, and many others. However, learning in these models typically converges to an equilibrium, either static or cyclic~\cite{czarnecki2020real}. They do not produce continual novelty in the data stream and thus cannot, on their own, produce compounding innovation. Below, we highlight two collective living processes that may contribute to compounding innovation: population pressures and arms races.

\subsection*{Population pressures}

Population growth in a resource-constrained environment leads to competition, where selection favors individuals who locate and exploit under-used resources. Adaptive dynamics models show that this selection can drive diversification in evolutionary strategies \cite{mcgill2007evolutionary}, as seen in adaptive radiations after colonisation events~\cite{schluter2000ecology}.

These ideas are used within genetics algorithms to maintain diversity in multi-objective optimisation~\cite{sareni1998fitness}. In multi-agent reinforcement learning, they led to Malthusian reinforcement learning~\cite{leibo2019malthusian} where sub-population sizes fluctuate based on their returns, thus creating local population pressures which drive agents away from sub-optimal equilibria. In artificial life and evolutionary computation, such competition has led to systems that surprised their creators with innovations exploiting misspecification of goals or errors in environment implementations~\cite{lehman2020surprising}.

Population pressures automatically motivate agents to seek new data. However, this search is not unbounded, as it depends on competition between agents; cooperative equilibria often remain unattainable.

\subsection*{Arms races}

Positive feedback loops in natural selection can give rise to evolutionary arms races~\cite{van1973new} which drive the evolution of extreme traits~\cite{dawkins1979arms}. For example, as cheetahs evolved to be faster to catch prey, they created selection pressure on their prey to be even faster, creating further selection on cheetahs. 

In multi-agent systems, arms races can provide autocurricula where agents continually adapt in response to one another's innovations, thus at any stage most interactions occur between agents with similar abilities. Consequently, the challenges confronted by agents scale with the abilities of surrounding agents, thereby providing gradients for learning otherwise unattainable optima (analogous to the cheetah-gazelle arms race). Within multi-agent RL, robotics, and artificial life, autocurricula using arms races have been shown to produce a succession of more and more complex skills~\cite{sims1994evolving, nolfi1998coevolving, Silver16Go, wang2019paired, baker2019emergent, stooke2021open, sukhbaatar2017intrinsic, johanson2022emergent, clune2019ai}. 

Autocurricula have been leveraged to create algorithms that co-adapt, such as dynamic environments posing learning challenges to agents~\cite{nisioti2020grounding, wang2019paired} and intrinsic motivation systems that generate their own goals~\cite{aubret2019survey}. Others learn through self-play to achieve a concrete objective, such as defeating a human expert at a game~\cite{tesauro1995td, Silver16Go, jaderberg2019human, meta2022human}. In these cases, training directly against the grandmaster would likely fail: the untrained agent would never win a game, so it would never get any reward signal to learn from. Multi-agent self-play allows na\"ive agents to discover strategies that work against na\"ive agents. As an agent's behaviour learning converges on a local optimum the distribution of data received by their opponents changes. Strategies that exploit known optima become increasingly rewarding, thereby motivating agents to explore new solutions.

Evidence suggests that innovation through arms races alone is limited. In two-player zero-sum games, as the skill of the agent increases, the space of possible policies collapses to a small set of intransitive strategies~\cite{czarnecki2020real}. Consequently, the novelty of the agent's experience stagnates. Compounding innovation appears to only be possible outside of the two-player zero-sum case, where further interactions may continue to enrich the agent's experience. In many cases, these interactions create incentives for social relationships.

\section*{Social relationships}

In addition to adapting to the general challenges of collective life, agents in persistent groups may gain advantage by attending to others as distinct individuals. These social pressures, and the cognitive machinery evolved to tackle them (Box~1), give rise to social dilemmas, situations where mutually beneficial interactions are possible despite risks. Of particular interest for collective intelligence is social learning, where individuals acquire new skills by imitating successful group members. When agents form social relationships this alters the data stream they encounter, creating new incentives for strategic communication and associated arms races. However, in the absence of other forces, these mechanisms can become trapped in new, collective equilibria. 

\subsection*{Machiavellian selection}

Cognition that incorporates awareness of group structure and relationships allows for division of labor, reciprocity and coalition formation, but also creates social dilemmas, as individuals can benefit from cooperating selectively. The social brain (or Machiavellian intelligence) hypothesis~\cite{dunbar1998social} proposes that the interplay of competitive and cooperative interactions arising in such relationships creates positive feedbacks in cognitive evolution. Theoretically, improved social cognition could enable an individual to better predict and control others, whilst making that individual harder to predict itself, thus creating a cognitive arms race~\cite{byrne1994machiavellian}. For example, long-lasting relationships create scope for complex cognitive strategies like reciprocity which in turn incentivise strategic cheating, cheater detection, and cheater detection evasion. Such social cognition autocurricula might lead to runaway cognitive evolution~\cite{byrne2018machiavellian, leibo2019autocurricula}. 

Theoretical results show that in populations playing two player zero-sum games, appropriate choices of which agent trains with which other agent, and who learns from that experience, can lead to optimal behaviour in the limit~\cite{lanctot2017unified}. The importance of the specific structure of interactions between individuals is further illustrated by Vinyals, et al.~\cite{vinyals2019grandmaster}. Building on the success of self-play models, AlphaStar extended multiplayer training with multiple distinct pools from which to select opposing teams, such that each pool enforced the learning of specific strategies. By training on multiple pools, agents remained robust to diverse opposing strategies, yielding performance rivaling that of champion StarCraft II players. Combining these techniques with a large language model to communicate intentionality via natural language led to human-level performance in \emph{Diplomacy}~\cite{meta2022human}.

\subsection*{Social learning}

Individuals may acquire new skills by imitating others' behaviour. In some species, such socially learned innovations can spread and persist over generations, forming cultural traditions. Animal traditions include songs, migration routes and foraging techniques in cetaceans and birds, and mate preferences and commuting routes in fish among others~\cite{whiten2019cultural}. In humans, social learning is thought to be the key factor in group cooperation. Yet, social learning may be self-limiting: when innovation is costly, imitation may out-compete it~\cite{rendell2010copy}, resulting in reduced innovation overall, which in turn decreases the value of imitation.

The collective outcomes of social learning are also highly sensitive to network topology, as illustrated by agent-based models of innovation~\cite{fang2010balancing, lazer2007network}. behavioural experiments show that topology influences diffusion of innovations~\cite{mason2008propagation}, propagation of beliefs~\cite{vlasceanu2021network}, and collective memory dynamics~\cite{coman2016mnemonic}. Optimal topology is task-dependent: decentralised networks appear most effective for groups facing collective problem-solving challenges~\cite{centola2022network}, whereas centralised topologies facilitate group innovation and collective action~\cite{bernstein2018intermittent, mckee2023scaffolding}. 

Because of its central role in human intelligence~\cite{boyd2011cultural}, social learning has attracted substantial attention from the AI community, with diverse techniques developed including imitation learning~\cite{osa2018algorithmic}, behavioural cloning from observation~\cite{torabi2018behavioral}, generative adversarial imitation learning~\cite{ho2016generative}, regularisation to human priors~\cite{liu2021motor}, and others. Social learning can also emerge spontaneously in RL agents. For example, they will readily learn to follow a more knowledgeable agent through a maze~\cite{borsa2019observational}. Such emergent social learning strategies can arise in novel environments and with unfamiliar experts~\cite{ndousse2021emergent}. Nisioti, et al.~\cite{nisioti2022social} argue that, like humans, RL agents can best discover hierarchical innovations when they share their experiences in dynamic communication topologies. These social learning algorithms aim to bootstrap the learning process of agents, so they need not learn everything on their own. Fine tuning of the socially learned behaviours is still done via reinforcement learning. However, social learning of this type does not lead to compounding innovation because the learned behaviours must already be in the data the agent learns from. 

Converging evidence suggests that human-level compounding innovation driven by social learning (i.e. cultural evolution) emerged from flexible multi-scale interactions arising within a confluence of population pressures, social relationships, and other evolutionary forces~\cite{jablonka2006evolution, henrich2016secret}. To use social learning effectively for ongoing data generation, we must first understand the dynamics underlying these multi-scale interactions.

\section*{Major transitions in evolution}

From genes upwards, the units of natural selection interact at multiple scales. Heterogeneous interactions among lower-level units create clusters that can be subject to natural selection in their own right, thus forming higher-level units. For lower-level units, selection often favors cooperation as their success is tied to that of the collective. In such groups, new mechanisms of information transmission may emerge to coordinate behaviour~\cite{jablonka2014evolution}. These mechanisms permit increasingly efficient self-control of the group, leading to the evolution of a new, group-level individuality. This event is termed a ``major evolutionary transition''~\cite{szathmary1995major}. The major transition creates a set of mutually reinforcing evolutionary trends, where selection favors coordinated cooperation, division of labour, and suppression or channelling of conflict. Examples of major transitions include the origin of chromosomes, endosymbiosis, multicellularity, nervous systems, eusocial animals, and cumulative culture in humans.

The major transitions paradigm offers insight into how incentives and interactions are dynamically rearranged by natural collectives to influence their internal adaptive landscape~\cite{jablonka2006evolution}. Information transmission allows the collective to shape the data streams of lower-level units via both competitive and cooperative incentives, outpacing natural selection to drive behaviour that serves its interests. For instance, cells in multicellular organisms manufacture an extracellular matrix which organises both mutual support and competition that suppresses harmful mutations and removes damaged cells~\cite{bowling2019cell}. During mammalian development, overproduction of cells is followed by competitive pruning, favoring cells that have formed appropriate social connections~\cite{raff1992social}. Competitive pheromone trails in ants and waggle dances in honey bees lead to rational collective foraging decisions based on individuals' private information.

Consider scenarios in which units face a social dilemma where unilateral deviation from selfishness is discouraged, but a joint deviation could escape a bad equilibrium's basin of attraction. If this escape benefits a higher-level unit, perhaps in competition with other higher-level units, then there is selective pressure to shape the data stream of the lower-level units such that they all experience a single binary choice where cooperation is preferred. This would cause all units to shift their strategy in a correlated manner, thus escaping the original dilemma. Similarly, the higher-level unit could bring about specialisation and division of labor by promoting competition and arms races among local clusters of units via appropriate manipulations of their data stream. 

In machine learning, the idea of using collectives with aligned incentives for problem solving has been explored from several angles. Manual alignment has enabled cooperation and coordination in a variety of contexts, including task partitioning~\cite{ferrante2015evolution}, and convention formation~\cite{peysakhovich2018prosocial}. Swarm robotics explores how local information and actions can lead to a desirable emergent global behaviour~\cite{brambilla2013swarm}, whilst cooperative multi-agent RL focuses on designing algorithms that improve learning of useful behaviours~\cite{oroojlooy2022review}. All of these approaches, however, tend to be brittle and generalise poorly to novel situations~\cite{schranz2020swarm, leibo2021scalable}. 

Hierarchical reinforcement learning (HRL) decomposes a problem into a hierarchical representation where low-level controllers implement policies at fine-grained scales (e.g. by solving sub-tasks), while high-level controllers decide which low-level controller to deploy, and for how long~\cite{sutton2018reinforcement}. Thus a collection of semi-independent policies is organised by a high level controller where the whole system is learned end-to-end. However, the interaction of learning scales between high- and low-level controllers poses challenges. For controllers to learn to deploy units effectively, the units must already be competent and specialised, but for units to learn useful policies, they require a learning signal of their usefulness for solving the high-level problem.

Recent work in AI has explored scenarios where alignment emerges during learning, for example by considering an agent composed of sub-agents which experience population pressures for credit assignment~\cite{sunehag2023diversity}, by evolving mixtures of incentives for cooperation~\cite{wang2019evolving}, or having a global agent shape incentives for a group of selfish learners~\cite{gemp2022d3c, zheng2022ai, koster2022human}. These approaches show that higher-level processes can generate incentive alignment at lower levels.

Insofar as coordination is achieved by manipulating the data experienced by low-level units, intelligence in low-level units is a two-edged sword: intelligent units may be more useful, but simpler units are more easily controlled. One effect repeatedly observed in major transitions is that lower-level units become simplified over time (see Box~1). This likely limits ongoing data generation---at least in cases other than cumulative culture.

\subsection*{Cumulative culture}

While there is evidence for cultural traditions in animals~\cite{whiten2019cultural}, humans appear uniquely adapted for accurate and flexible information transmission using language and active teaching~\cite{dean2012identification}, such that learned information accumulates across multiple generations (cumulative culture). Cumulative culture is the mechanism behind the compounding innovation of human technology and knowledge.

Herrmann, et al.~\cite{herrmann2007humans} argued that extended social learning provides an impetus for cognitive flexibility as the suite of learned skills in a population expands and is refined across generations. This self-reinforcing cycle may account simultaneously for increased brain size, cognitive flexibility, division of labor and technological evolution in humans~\cite{muthukrishna2016innovation}. Compositional language, where new meaningful expressions can be constructed from existing ones, is key to this process~\cite{heyes2018cognitive}. Before the origin of language, autocurricula based in Machiavellian or social learning arms races, whilst in principle potentially runaway processes, were subject to physical and bioenergetic constraints on brain size~\cite{dunbar2017there}. Language relaxed these constraints by allowing efficiently distributed data representation and storage via oral tradition~\cite{kirby2015compression}.

Efficient information transmission via language allows behavioural adaptation of groups to outpace natural selection. Thus the origin of language arguably constitutes a major evolutionary transition. However, unlike the transition to multicellularity, the cultural transition has not yet led to the emergence of discrete super-organisms. Instead, social groups and individuals coexist, with individuals moving freely between overlapping groups. Language permits evolution of cultural structures that manipulate human data streams to dynamically manage human incentives, allowing groups to flexibly reorganise social network topology to adapt to current problems, generate cooperation and channel competition through norms and institutions for group-level benefits. Institutions can organise diverse individual motivations to align them with societal needs, for example through storytelling, reputation, punishment, and economic incentives~\cite{ostrom2005understanding}. In this way, human societies can harness self-motivated individual brains to excel at diverse specialised tasks without compromising generality at cultural scales.

Can AI achieve something similar? Research has shown that symbolic communication can emerge in multi-agent settings given appropriate training pressures~\cite{lazaridou2016multi, havrylov2017emergence, mordatch2018emergence}. More recently large-scale language models including GPT-3~\cite{brown2020language} and PaLM~\cite{chowdhery2022palm} show that training to generate human-like natural language at scale yields the capacity to converse on diverse topics and adapt to novel tasks from very small numbers of examples~\cite{chan2022data, wei2022chain}. It has been argued that such models may have difficulty consistently associating words with their meanings in the physical and/or social world~\cite{bisk2020experience, ullman2023large}. However, there are promising ways to alleviate this issue, for example by using multi-modal datasets to associate concepts across modalities~\cite{alayrac2022flamingo}, connecting language models to simulators of the physical world~\cite{liu2022mind}, and by giving them access to internet search engines~\cite{glaese2022improving}. Some recent work has proposed using language directly as a learning environment to imbue a solipsistic AI agent with its own goals~\cite{colas2022language}.

Large language models are already being used to produce large amounts of content on the internet. For humans, our ongoing cultural data generation has been driven by autocurricula underpinned by population pressures, evolutionary arms races and social relationships, and channeled by multi-scale cultural selection to allow flexible and dynamic cooperative division of labor along with specialised skills. Perhaps analogous processes have already begun to play out on the internet, with humans and AIs generating a data stream conducive to AI cumulative culture. However, early results show that training these models on their own outputs might lead to their collapse, losing more of the tails of the data distribution with each successive retraining~\cite{shumailov2023curse}. It remains critical to consider the source and dynamics of their data stream.

\section*{Synthesis}

Many contemporary authors argue for the importance of a data-rich environment to learning systems, and our most successful current algorithms effectively leverage the accumulated informational output of humanity. Indeed, it has been argued that achieving artificial general intelligence is simply a matter of consuming more such data. But assimilation does not imply creation. As large language models outpace human data generation, the collective output of human culture may become the bottleneck for AI~\cite{villalobos2022will}.

Insights from evolution suggest a way forward: ongoing data generation may arise from structured interactions of agents, as they compete and cooperate in populations. Specific interaction structures generate autocurricula, including social relationships, multi-scale coordination via social learning, division of labor, and cumulative cultural transmission. In biology, these processes continually modulate the data streams received by individuals. Population pressures reduce the attraction of well-known optima, reducing the likelihood of convergence to a stationary or periodic data stream. Arms races refine the experiences pertinent to specific skills. Social relationships allow individuals to cooperate, negotiate and learn from one another, disseminating innovations and exposing cooperative equilibria inaccessible to solipsistic agents. Major transitions generate coordinated super-organisms where conflict and cooperation among lower-level units are organised through data control by higher-level units, leading to efficient division of labor among lower-level units. Finally, compositional language allows humans to rapidly manipulate and share data streams in dynamically structured social networks, providing opportunities for flexible learning and multi-scale division of labor unprecedented in evolutionary history. Although there is some overlap between these mechanisms, each is punctuated by a specific type of interaction that influences the data stream available for learning in qualitatively different ways.

Individually each of the mechanisms discussed in this perspective has yielded only partial success in the quest for an artificial system capable of compounding innovation. We posit that a system that combines all of them is more likely to succeed. But how do we combine them? One could design a system in which all of those mechanisms emerge from first principles, or one could digitise the main aspects of each of them, hard-coding them into a system. We think the right approach is somewhere in between. There are some aspects for which emergence and flexibility are key, like the flexible incentive alignment and data manipulation of major transitions and cumulative culture. Yet, there are others for which an engineered solution might be adequate, like providing agents with a curiosity or a social learning module.

Recent decades have seen a Cambrian explosion of learning algorithms, many of them inspired by living systems. Within machine learning, the groundwork for representing these processes has already been laid in diverse algorithmic methodologies. We suggest an approach that integrates not only the components of AI, but its underlying generative processes, in particular the mechanisms that lead to compounding innovation by enriching data streams. 

\noindent\fbox{%
    \parbox{\textwidth}{%

\textbf{Biological underpinnings of relationships and major transitions}

\label{box:relationships}
Social relationships between individuals occur in diverse taxa, but are best known in vertebrates. Group hunters such as lions and bottlenose dolphins exploit the collective vigilance of prey to drive them into coordinated ambushes~\cite{gazda2016driver}. Pair-bonded species may adopt sex-specific roles in parental care, or use a temporal division of labour, or (as in the case of snapping shrimp) cooperate without parenting~\cite{bales2021pair}. Reciprocity and coalitions arise in persistent social groups of unrelated individuals, whilst alloparenting and division of labour more often emerge when relatedness is high~\cite{lukas2018social}. From a game theoretic perspective, these behaviours imply diverse iterated multi-player social dilemmas that support a rich strategic space.

A key aspect of relationships is social bonding, the formation of preferential attachments between individuals. Far from learning to solve social dilemmas from a blank slate, cooperating individuals rely on pre-wired neuroendocrine networks to organise bonded relationships. Social bonds reduce cognitive load in managing relationships, and thus provide an evolutionary foundation for the evolution of the social brain.

In mammals, for example, lactation makes the mother-offspring relationship critical to fitness. Oxytocin surges during birth and lactation trigger release of dopamine and endogenous opioids that facilitate learning, attachment and sensitive communication between mother and offspring. Additional oxytocin release occurs throughout infancy during behaviours such as touch, vocalisation, and mutual gaze, reinforcing these patterns. These behaviours and neuroendocrine feedbacks provide an evolutionary template for other mammalian social bonds that are predominantly prosocial~\cite{feldman2015adaptive}. Pair bonds, friendships and group cooperation systems arise as extensions of this template that use behavioural triggers (such as eye contact, grooming, vocalisations and coordinated dancing~\cite{tarr2015synchrony}) to stimulate social bonding neurochemistry~\cite{lieberwirth2014social}. Male chimpanzees, for example, establish and maintain coalitions through grooming; coalitions are a key factor in hierarchies which strongly influence mating success and thus fitness. 

As social groups approach a major transition, individual cognitive sophistication may decline because the emerging higher-level unit evolves to regulate these feedbacks. When division of labor arises, general cognitive ability of underlying units may create inefficiency or even pose a threat to collective control. Selfish genetic elements in genomes, cancer in multicellular organisms, and worker reproduction in eusocial superorganisms represent classic examples where innovation at smaller scales threatens a larger-scale entity, driving costly policing adaptations~\cite{aagren2019enforcement}. This leads to a pattern evident across scales where units become simplified and interdependent. Mitochondria and chloroplasts come to depend on nuclear genes, cells that form interdependent tissues and organs become incapable of survival alone, and eusocial insects form specialised castes with reduced brain size. 

The same trends occur in social mammals: as kinship increases within groups, division of labor, alloparenting and reproductive suppression emerge, and aggression, reciprocity and coalition formation decline along with brain and neocortex size~\cite{lukas2018social}. Among mammals domesticated by humans (and possibly including humans), a syndrome has been noted in which evolution retards neural crest developmental pathways regulating brain size, facial morphology and aggression, creating a prolonged juvenile period and small-brained, friendly, neotenous adult~\cite{wilkins2014domestication}. Thus the major transition, whilst opening up new possibilities for collective cognition, may be ultimately self-limiting in data generation due to the control needs of the unified collective. 
    }%
}

\section*{Acknowledgments}
We thank Ankesh Anand, David Parkes, Tom Schaul, and Karl Tuyls for helpful comments on early versions of this manuscript.

\section*{Competing interests}
The authors declare no competing interests

\section*{Additional information}

\paragraph{Author contributions} All authors contributed ideas and wrote the paper.


\begin{thebibliography}{100}
\providecommand{\bibinfo}[2]{#2}

\bibitem{krizhevsky2012imagenet}
\bibinfo{author}{Krizhevsky, A.}, \bibinfo{author}{Sutskever, I.} \&
  \bibinfo{author}{Hinton, G.E.}
\newblock \bibinfo{title}{Imagenet classification with deep convolutional
  neural networks}.
\newblock \emph{\bibinfo{journal}{Adv. {NeurIPS}}}
  \textbf{\bibinfo{volume}{25}} (\bibinfo{year}{2012}).

\bibitem{deng2009imagenet}
\bibinfo{author}{Deng, J.} \emph{et~al.}
\newblock \bibinfo{title}{Imagenet: A large-scale hierarchical image database}.
\newblock \emph{\bibinfo{journal}{{IEEE Conf. Comput. Vis. Pattern Recog.}}}
  \bibinfo{pages}{248--255} (\bibinfo{year}{2009}).

\bibitem{kaplan2020scaling}
\bibinfo{author}{Kaplan, J.} \emph{et~al.}
\newblock \bibinfo{title}{Scaling laws for neural language models}.
\newblock {\bibinfo{journal}{Preprint at https://arXiv.org/abs/2001.08361}}
  (\bibinfo{year}{2020}).

\bibitem{bommasani2021opportunities}
\bibinfo{author}{Bommasani, R.} \emph{et~al.}
\newblock \bibinfo{title}{On the opportunities and risks of foundation models}.
\newblock {\bibinfo{journal}{Preprint at https://arXiv.org/abs/2108.07258}}
  (\bibinfo{year}{2021}).

\bibitem{hoffmann2022training}
\bibinfo{author}{Hoffmann, J.} \emph{et~al.}
\newblock \bibinfo{title}{Training compute-optimal large language models}.
\newblock {\bibinfo{journal}{Preprint at https://arXiv.org/abs/2203.15556}}
  (\bibinfo{year}{2022}).

\bibitem{fei2022searching}
\bibinfo{author}{Fei-Fei, L.} \& \bibinfo{author}{Krishna, R.}
\newblock \bibinfo{title}{Searching for computer vision north stars}.
\newblock \emph{\bibinfo{journal}{Daedalus}} \textbf{\bibinfo{volume}{151}},
  \bibinfo{pages}{85--99} (\bibinfo{year}{2022}).

\bibitem{alayrac2022flamingo}
\bibinfo{author}{Alayrac, J.-B.} \emph{et~al.}
\newblock \bibinfo{title}{Flamingo: A visual language model for few-shot
  learning}.
\newblock \emph{\bibinfo{journal}{Adv. {NeurIPS}}}
  \textbf{\bibinfo{volume}{35}} \bibinfo{pages}{23716-23736} (\bibinfo{year}{2022}).

\bibitem{young1804bakerian}
\bibinfo{author}{Young, T.}
\newblock \bibinfo{title}{I. The Bakerian Lecture. Experiments and calculations
  relative to physical optics}.
\newblock \emph{\bibinfo{journal}{Philos. Trans. Royal Soc. B}}
  \bibinfo{pages}{1--16} (\bibinfo{year}{1804}).

\bibitem{sutton2018reinforcement}
\bibinfo{author}{Sutton, R.S.} \& \bibinfo{author}{Barto, A.G.}
\newblock \emph{\bibinfo{title}{Reinforcement learning: An introduction}}
  (\bibinfo{publisher}{{MIT Press}}, \bibinfo{year}{2018}).

\bibitem{schaul2019ray}
\bibinfo{author}{Schaul, T.}, \bibinfo{author}{Borsa, D.},
  \bibinfo{author}{Modayil, J.} \& \bibinfo{author}{Pascanu, R.}
\newblock \bibinfo{title}{Ray interference: A source of plateaus in deep
  reinforcement learning}.
\newblock {\bibinfo{journal}{Preprint at https://arXiv.org/abs/1904.11455}}
  (\bibinfo{year}{2019}).

\bibitem{ortega2021shaking}
\bibinfo{author}{Ortega, P.A.} \emph{et~al.}
\newblock \bibinfo{title}{Shaking the foundations: Delusions in sequence models
  for interaction and control}.
\newblock {\bibinfo{journal}{Preprint at https://arXiv.org/abs/2110.10819}}
  (\bibinfo{year}{2021}).

\bibitem{huang2022large}
\bibinfo{author}{Huang, J.} \emph{et~al.}
\newblock \bibinfo{title}{Large language models can self-improve}.
\newblock {\bibinfo{journal}{Preprint at https://arXiv.org/abs/2210.11610}}
  (\bibinfo{year}{2022}).

\bibitem{shumailov2023curse}
\bibinfo{author}{Shumailov, I.} \emph{et~al.}
\newblock \bibinfo{title}{The curse of recursion: Training on generated data
  makes models forget}.
\newblock {\bibinfo{journal}{Preprint at https://arXiv.org/abs/2305.17493}}
  (\bibinfo{year}{2023}).

\bibitem{wang2019paired}
\bibinfo{author}{Wang, R.}, \bibinfo{author}{Lehman, J.},
  \bibinfo{author}{Clune, J.} \& \bibinfo{author}{Stanley, K.O.}
\newblock \bibinfo{title}{Paired open-ended trailblazer ({POET}): Endlessly
  generating increasingly complex and diverse learning environments and their
  solutions}.
\newblock {\bibinfo{journal}{Preprint at https://arXiv.org/abs/1901.01753}}
  (\bibinfo{year}{2019}).

\bibitem{Portelas2020automatic}
\bibinfo{author}{Portelas, R.}, \bibinfo{author}{Colas, C.},
  \bibinfo{author}{Weng, L.}, \bibinfo{author}{Hofmann, K.} \&
  \bibinfo{author}{Oudeyer, P.-Y.}
\newblock \bibinfo{title}{Automatic curriculum learning for Deep RL: A short
  survey}.
\newblock \emph{\bibinfo{journal}{Proc. {IJCAI}}} \bibinfo{pages}{4819--4825}
  (\bibinfo{year}{2020}).
\newblock \bibinfo{note}{Survey track}.

\bibitem{linke2020adapting}
\bibinfo{author}{Linke, C.}, \bibinfo{author}{Ady, N.M.},
  \bibinfo{author}{White, M.}, \bibinfo{author}{Degris, T.} \&
  \bibinfo{author}{White, A.}
\newblock \bibinfo{title}{Adapting behavior via intrinsic reward: A survey and
  empirical study}.
\newblock \emph{\bibinfo{journal}{J Artif. Intell. Res.}}
  \textbf{\bibinfo{volume}{69}}, \bibinfo{pages}{1287--1332}
  (\bibinfo{year}{2020}).

\bibitem{oudeyer2007intrinsic}
\bibinfo{author}{Oudeyer, P.-Y.} \& \bibinfo{author}{Kaplan, F.}
\newblock \bibinfo{title}{What is intrinsic motivation? a typology of
  computational approaches}.
\newblock \emph{\bibinfo{journal}{Front. Neurorobot.}}
  \textbf{\bibinfo{volume}{1}}, \bibinfo{pages}{6} (\bibinfo{year}{2007}).

\bibitem{pathak2017curiosity}
\bibinfo{author}{Pathak, D.}, \bibinfo{author}{Agrawal, P.},
  \bibinfo{author}{Efros, A.A.} \& \bibinfo{author}{Darrell, T.}
\newblock \bibinfo{title}{Curiosity-driven exploration by self-supervised
  prediction}.
\newblock \emph{\bibinfo{journal}{Proc. {ICML}}} \bibinfo{pages}{2778--2787}
  (\bibinfo{year}{2017}).

\bibitem{colas2022autotelic}
\bibinfo{author}{Colas, C.}, \bibinfo{author}{Karch, T.},
  \bibinfo{author}{Sigaud, O.} \& \bibinfo{author}{Oudeyer, P.-Y.}
\newblock \bibinfo{title}{Autotelic agents with intrinsically motivated
  goal-conditioned reinforcement learning: A short survey}.
\newblock \emph{\bibinfo{journal}{J. Artif. Intell. Res.}}
  \textbf{\bibinfo{volume}{74}}, \bibinfo{pages}{1159--1199}
  (\bibinfo{year}{2022}).

\bibitem{ladosz2022exploration}
\bibinfo{author}{Ladosz, P.}, \bibinfo{author}{Weng, L.}, \bibinfo{author}{Kim,
  M.} \& \bibinfo{author}{Oh, H.}
\newblock \bibinfo{title}{Exploration in deep reinforcement learning: A
  survey}.
\newblock \emph{\bibinfo{journal}{Inf. Fusion}}  (\bibinfo{year}{2022}).

\bibitem{jiang2023general}
\bibinfo{author}{Jiang, M.}, \bibinfo{author}{Rockt{\"a}schel, T.} \&
  \bibinfo{author}{Grefenstette, E.}
\newblock \bibinfo{title}{General intelligence requires rethinking
  exploration}.
\newblock \emph{\bibinfo{journal}{Royal Soc. Open Sci.}}
  \textbf{\bibinfo{volume}{10}}, \bibinfo{pages}{230539}
  (\bibinfo{year}{2023}).

\bibitem{kearns2002near}
\bibinfo{author}{Kearns, M.} \& \bibinfo{author}{Singh, S.}
\newblock \bibinfo{title}{Near-optimal reinforcement learning in polynomial
  time}.
\newblock \emph{\bibinfo{journal}{Mach. Learn.}} \textbf{\bibinfo{volume}{49}},
  \bibinfo{pages}{209--232} (\bibinfo{year}{2002}).

\bibitem{osband2019deep}
\bibinfo{author}{Osband, I.}, \bibinfo{author}{Van~Roy, B.},
  \bibinfo{author}{Russo, D.J.} \& \bibinfo{author}{Wen, Z.}
\newblock \bibinfo{title}{Deep exploration via randomized value functions.}
\newblock \emph{\bibinfo{journal}{J. Mach. Learn. Res.}}
  \textbf{\bibinfo{volume}{20}}, \bibinfo{pages}{1--62} (\bibinfo{year}{2019}).

\bibitem{leibo2019autocurricula}
\bibinfo{author}{Leibo, J.Z.}, \bibinfo{author}{Hughes, E.},
  \bibinfo{author}{Lanctot, M.} \& \bibinfo{author}{Graepel, T.}
\newblock \bibinfo{title}{Autocurricula and the emergence of innovation from
  social interaction: A manifesto for multi-agent intelligence research}.
\newblock {\bibinfo{journal}{Preprint at https://arXiv.org/abs/1903.00742}}
  (\bibinfo{year}{2019}).

\bibitem{sukhbaatar2017intrinsic}
\bibinfo{author}{Sukhbaatar, S.} \emph{et~al.}
\newblock \bibinfo{title}{Intrinsic motivation and automatic curricula via
  asymmetric self-play}.
\newblock \emph{\bibinfo{journal}{Proc. {ICLR}}}
  (\bibinfo{year}{2018}).

\bibitem{leibo2019malthusian}
\bibinfo{author}{Leibo, J.Z.} \emph{et~al.}
\newblock \bibinfo{title}{Malthusian reinforcement learning}.
\newblock \emph{\bibinfo{journal}{Proc. {AAMAS}}} \bibinfo{pages}{1099--1107}
  (\bibinfo{year}{2019}).

\bibitem{baker2019emergent}
\bibinfo{author}{Baker, B.} \emph{et~al.}
\newblock \bibinfo{title}{Emergent tool use from multi-agent autocurricula}.
\newblock \emph{\bibinfo{journal}{Proc. {ICLR}}}
  (\bibinfo{year}{2020}).

\bibitem{balduzzi2019open}
\bibinfo{author}{Balduzzi, D.} \emph{et~al.}
\newblock \bibinfo{title}{Open-ended learning in symmetric zero-sum games}.
\newblock \emph{\bibinfo{journal}{Proc. ICML}} \bibinfo{pages}{434--443}
  (\bibinfo{year}{2019}).

\bibitem{openai2021asymmetric}
\bibinfo{author}{Plappert, M.} \emph{et~al.}
\newblock \bibinfo{title}{Asymmetric self-play for automatic goal discovery in
  robotic manipulation}.
\newblock {\bibinfo{journal}{Preprint at https://arXiv.org/abs/2101.04882}}
  (\bibinfo{year}{2021}).

\bibitem{goodfellow2014generative}
\bibinfo{author}{Goodfellow, I.} \emph{et~al.}
\newblock \bibinfo{title}{Generative adversarial nets}.
\newblock \emph{\bibinfo{journal}{Adv. {NeurIPS}}}
  \textbf{\bibinfo{volume}{27}} (\bibinfo{year}{2014}).

\bibitem{herrmann2007humans}
\bibinfo{author}{Herrmann, E.}, \bibinfo{author}{Call, J.},
  \bibinfo{author}{Hern{\'a}ndez-Lloreda, M.V.}, \bibinfo{author}{Hare, B.} \&
  \bibinfo{author}{Tomasello, M.}
\newblock \bibinfo{title}{Humans have evolved specialized skills of social
  cognition: The cultural intelligence hypothesis}.
\newblock \emph{\bibinfo{journal}{Science}} \textbf{\bibinfo{volume}{317}},
  \bibinfo{pages}{1360--1366} (\bibinfo{year}{2007}).

\bibitem{boyd2011cultural}
\bibinfo{author}{Boyd, R.}, \bibinfo{author}{Richerson, P.J.} \&
  \bibinfo{author}{Henrich, J.}
\newblock \bibinfo{title}{The cultural niche: Why social learning is essential
  for human adaptation}.
\newblock \emph{\bibinfo{journal}{Proc. Natl. Acad. Sci.}}
  \textbf{\bibinfo{volume}{108}}, \bibinfo{pages}{10918--10925}
  (\bibinfo{year}{2011}).

\bibitem{whiten2019cultural}
\bibinfo{author}{Whiten, A.}
\newblock \bibinfo{title}{Cultural evolution in animals}.
\newblock \emph{\bibinfo{journal}{Annu. Rev. Ecol. Evol. Syst.}}
  \textbf{\bibinfo{volume}{50}}, \bibinfo{pages}{27--48}
  (\bibinfo{year}{2019}).

\bibitem{dunbar1998social}
\bibinfo{author}{Dunbar, R. I.M.}
\newblock \bibinfo{title}{The social brain hypothesis}.
\newblock \emph{\bibinfo{journal}{Evol. Anthropol.}}
  \textbf{\bibinfo{volume}{6}}, \bibinfo{pages}{178--190}
  (\bibinfo{year}{1998}).

\bibitem{byrne2018machiavellian}
\bibinfo{author}{Byrne, R.W.}
\newblock \bibinfo{title}{Machiavellian intelligence retrospective.}
\newblock \emph{\bibinfo{journal}{J. Comp. Psychol.}}
  \textbf{\bibinfo{volume}{132}}, \bibinfo{pages}{432} (\bibinfo{year}{2018}).

\bibitem{szathmary1995major}
\bibinfo{author}{Szathm{\'a}ry, E.} \& \bibinfo{author}{Maynard~Smith, J.}
\newblock \bibinfo{title}{The major evolutionary transitions}.
\newblock \emph{\bibinfo{journal}{Nature}} \textbf{\bibinfo{volume}{374}},
  \bibinfo{pages}{227--232} (\bibinfo{year}{1995}).

\bibitem{jablonka2014evolution}
\bibinfo{author}{Jablonka, E.} \& \bibinfo{author}{Lamb, M.J.}
\newblock \emph{\bibinfo{title}{Evolution in four dimensions: Genetic,
  epigenetic, behavioral, and symbolic variation in the history of life}}
  (\bibinfo{publisher}{{MIT Press}}, \bibinfo{year}{2014}).

\bibitem{heyes2018cognitive}
\bibinfo{author}{Heyes, C.}
\newblock \emph{\bibinfo{title}{Cognitive gadgets: The cultural evolution of
  thinking}}  (\bibinfo{publisher}{Harvard Univ. Press},
  \bibinfo{year}{2018}).

\bibitem{ng2009bacterial}
\bibinfo{author}{Ng, W.-L.} \& \bibinfo{author}{Bassler, B.L.}
\newblock \bibinfo{title}{Bacterial quorum-sensing network architectures}.
\newblock \emph{\bibinfo{journal}{Ann. Rev. Genet.}}
  \textbf{\bibinfo{volume}{43}}, \bibinfo{pages}{197} (\bibinfo{year}{2009}).

\bibitem{verheggen2010alarm}
\bibinfo{author}{Verheggen, F.J.}, \bibinfo{author}{Haubruge, E.} \&
  \bibinfo{author}{Mescher, M.C.}
\newblock \bibinfo{title}{Alarm pheromones—chemical signaling in response to
  danger}.
\newblock \emph{\bibinfo{journal}{Vit. Horm.}} \textbf{\bibinfo{volume}{83}},
  \bibinfo{pages}{215--239} (\bibinfo{year}{2010}).

\bibitem{nagy2020synergistic}
\bibinfo{author}{Nagy, M.} \emph{et~al.}
\newblock \bibinfo{title}{Synergistic benefits of group search in rats}.
\newblock \emph{\bibinfo{journal}{Curr. Biol.}} \textbf{\bibinfo{volume}{30}},
  \bibinfo{pages}{4733--4738} (\bibinfo{year}{2020}).

\bibitem{schluter2000ecology}
\bibinfo{author}{Schluter, D.}
\newblock \emph{\bibinfo{title}{The ecology of adaptive radiation}}
  (\bibinfo{publisher}{Oxford Univ. Press}, \bibinfo{year}{2000}).

\bibitem{bansal2017emergent}
\bibinfo{author}{Bansal, T.}, \bibinfo{author}{Pachocki, J.},
  \bibinfo{author}{Sidor, S.}, \bibinfo{author}{Sutskever, I.} \&
  \bibinfo{author}{Mordatch, I.}
\newblock \bibinfo{title}{Emergent complexity via multi-agent competition}.
\newblock \emph{\bibinfo{journal}{Proc. {ICLR}}}
  (\bibinfo{year}{2018}).

\bibitem{reynolds1987flocks}
\bibinfo{author}{Reynolds, C.W.}
\newblock \bibinfo{title}{Flocks, herds and schools: A distributed behavioral
  model}.
\newblock \emph{\bibinfo{journal}{Proc. Ann. Conf. Comp. Graph. Interact.
  Tech.}} \bibinfo{pages}{25--34} (\bibinfo{year}{1987}).

\bibitem{lerer2017maintaining}
\bibinfo{author}{Lerer, A.} \& \bibinfo{author}{Peysakhovich, A.}
\newblock \bibinfo{title}{Maintaining cooperation in complex social dilemmas
  using deep reinforcement learning}.
\newblock {\bibinfo{journal}{Preprint at https://arXiv.org/abs/1707.01068}}
  (\bibinfo{year}{2017}).

\bibitem{leibo2017multiagent}
\bibinfo{author}{Leibo, J.Z.}, \bibinfo{author}{Zambaldi, V.},
  \bibinfo{author}{Lanctot, M.}, \bibinfo{author}{Marecki, J.} \&
  \bibinfo{author}{Graepel, T.}
\newblock \bibinfo{title}{{Multi-agent Reinforcement Learning in Sequential
  Social Dilemmas}}.
\newblock \emph{\bibinfo{journal}{Proc. {AAMAS}}}  (\bibinfo{year}{2017}).

\bibitem{mckee2022quantifying}
\bibinfo{author}{McKee, K.R.}, \bibinfo{author}{Leibo, J.Z.},
  \bibinfo{author}{Beattie, C.} \& \bibinfo{author}{Everett, R.}
\newblock \bibinfo{title}{Quantifying the effects of environment and population
  diversity in multi-agent reinforcement learning}.
\newblock \emph{\bibinfo{journal}{Auton. Agents Multi-Agent Syst.}}
  \textbf{\bibinfo{volume}{36}} (\bibinfo{year}{2022}).

\bibitem{strouse2021collaborating}
\bibinfo{author}{Strouse, D.}, \bibinfo{author}{McKee, K.},
  \bibinfo{author}{Botvinick, M.}, \bibinfo{author}{Hughes, E.} \&
  \bibinfo{author}{Everett, R.}
\newblock \bibinfo{title}{Collaborating with humans without human data}.
\newblock \emph{\bibinfo{journal}{Adv. {NeurIPS}}}
  \textbf{\bibinfo{volume}{34}}, \bibinfo{pages}{14502--14515}
  (\bibinfo{year}{2021}).

\bibitem{lazaridou2016multi}
\bibinfo{author}{Lazaridou, A.}, \bibinfo{author}{Peysakhovich, A.} \&
  \bibinfo{author}{Baroni, M.}
\newblock \bibinfo{title}{Multi-agent cooperation and the emergence of
  (natural) language}.
\newblock \emph{\bibinfo{journal}{Proc. {ICLR}}}
  (\bibinfo{year}{2017}).

\bibitem{czarnecki2020real}
\bibinfo{author}{Czarnecki, W.M.} \emph{et~al.}
\newblock \bibinfo{title}{Real world games look like spinning tops}.
\newblock \emph{\bibinfo{journal}{Adv. {NeurIPS}}}
  \textbf{\bibinfo{volume}{33}}, \bibinfo{pages}{17443--17454}
  (\bibinfo{year}{2020}).

\bibitem{mcgill2007evolutionary}
\bibinfo{author}{McGill, B.J.} \& \bibinfo{author}{Brown, J.S.}
\newblock \bibinfo{title}{Evolutionary game theory and adaptive dynamics of
  continuous traits}.
\newblock \emph{\bibinfo{journal}{Annu. Rev. Ecol. Evol. Syst.}}
  \textbf{\bibinfo{volume}{38}}, \bibinfo{pages}{403--435}
  (\bibinfo{year}{2007}).

\bibitem{sareni1998fitness}
\bibinfo{author}{Sareni, B.} \& \bibinfo{author}{Krahenbuhl, L.}
\newblock \bibinfo{title}{Fitness sharing and niching methods revisited}.
\newblock \emph{\bibinfo{journal}{IEEE Trans. Evol. Comp.}}
  \textbf{\bibinfo{volume}{2}}, \bibinfo{pages}{97--106}
  (\bibinfo{year}{1998}).

\bibitem{lehman2020surprising}
\bibinfo{author}{Lehman, J.} \emph{et~al.}
\newblock \bibinfo{title}{The surprising creativity of digital evolution: A
  collection of anecdotes from the evolutionary computation and artificial life
  research communities}.
\newblock \emph{\bibinfo{journal}{Artif. Life}} \textbf{\bibinfo{volume}{26}},
  \bibinfo{pages}{274--306} (\bibinfo{year}{2020}).

\bibitem{van1973new}
\bibinfo{author}{Van~Valen, L.}
\newblock \bibinfo{title}{A new evolutionary law}.
\newblock \emph{\bibinfo{journal}{Evol. Theory}} \textbf{\bibinfo{volume}{1}},
  \bibinfo{pages}{1--30} (\bibinfo{year}{1973}).

\bibitem{dawkins1979arms}
\bibinfo{author}{Dawkins, R.} \& \bibinfo{author}{Krebs, J.R.}
\newblock \bibinfo{title}{Arms races between and within species}.
\newblock \emph{\bibinfo{journal}{Proc. Royal Soc. B}}
  \textbf{\bibinfo{volume}{205}}, \bibinfo{pages}{489--511}
  (\bibinfo{year}{1979}).

\bibitem{sims1994evolving}
\bibinfo{author}{Sims, K.}
\newblock \bibinfo{title}{Evolving 3D morphology and behavior by competition}.
\newblock \emph{\bibinfo{journal}{Artif. Life}} \textbf{\bibinfo{volume}{1}},
  \bibinfo{pages}{353--372} (\bibinfo{year}{1994}).

\bibitem{nolfi1998coevolving}
\bibinfo{author}{Nolfi, S.} \& \bibinfo{author}{Floreano, D.}
\newblock \bibinfo{title}{Coevolving predator and prey robots: Do ``arms
  races'' arise in artificial evolution?}
\newblock \emph{\bibinfo{journal}{Artif. Life}} \textbf{\bibinfo{volume}{4}},
  \bibinfo{pages}{311--335} (\bibinfo{year}{1998}).

\bibitem{Silver16Go}
\bibinfo{author}{Silver, D.} \emph{et~al.}
\newblock \bibinfo{title}{Mastering the game of {G}o with deep neural networks
  and tree search.}
\newblock \emph{\bibinfo{journal}{Nature}} \textbf{\bibinfo{volume}{529}},
  \bibinfo{pages}{484–489} (\bibinfo{year}{2016}).

\bibitem{stooke2021open}
\bibinfo{author}{Stooke, A.} \emph{et~al.}
\newblock \bibinfo{title}{Open-ended learning leads to generally capable
  agents}.
\newblock {\bibinfo{journal}{Preprint at https://arXiv.org/abs/2107.12808}}
  (\bibinfo{year}{2021}).

\bibitem{johanson2022emergent}
\bibinfo{author}{Johanson, M.B.}, \bibinfo{author}{Hughes, E.},
  \bibinfo{author}{Timbers, F.} \& \bibinfo{author}{Leibo, J.Z.}
\newblock \bibinfo{title}{Emergent bartering behaviour in multi-agent
  reinforcement learning}.
\newblock {\bibinfo{journal}{Preprint at https://arXiv.org/abs/2205.06760}}
  (\bibinfo{year}{2022}).

\bibitem{clune2019ai}
\bibinfo{author}{Clune, J.}
\newblock \bibinfo{title}{{AI-GA}s: {AI}-generating algorithms, an alternate
  paradigm for producing general artificial intelligence}.
\newblock {\bibinfo{journal}{Preprint at https://arXiv.org/abs/1905.10985}}
  (\bibinfo{year}{2019}).

\bibitem{nisioti2020grounding}
\bibinfo{author}{Nisioti, E.} \& \bibinfo{author}{Moulin-Frier, C.}
\newblock \bibinfo{title}{Grounding artificial intelligence in the origins of
  human behavior}.
\newblock {\bibinfo{journal}{Preprint at https://arXiv.org/abs/2012.08564}}
  (\bibinfo{year}{2020}).

\bibitem{aubret2019survey}
\bibinfo{author}{Aubret, A.}, \bibinfo{author}{Matignon, L.} \&
  \bibinfo{author}{Hassas, S.}
\newblock \bibinfo{title}{A survey on intrinsic motivation in reinforcement
  learning}.
\newblock {\bibinfo{journal}{Preprint at https://arXiv.org/abs/1908.06976}}
  (\bibinfo{year}{2019}).

\bibitem{tesauro1995td}
\bibinfo{author}{Tesauro, G.}
\newblock \bibinfo{title}{ in \textit{{TD}-gammon, a self-teaching backgammon
  program, achieves master-level play}}  \bibinfo{pages}{267--285}
  (\bibinfo{publisher}{Springer}, \bibinfo{year}{1995}).

\bibitem{jaderberg2019human}
\bibinfo{author}{Jaderberg, M.} \emph{et~al.}
\newblock \bibinfo{title}{Human-level performance in {3D} multiplayer games
  with population-based reinforcement learning}.
\newblock \emph{\bibinfo{journal}{Science}} \textbf{\bibinfo{volume}{364}},
  \bibinfo{pages}{859--865} (\bibinfo{year}{2019}).

\bibitem{meta2022human}
\bibinfo{author}{Bakhtin, A.} \emph{et~al.}
\newblock \bibinfo{title}{Human-level play in the game of diplomacy by
  combining language models with strategic reasoning}.
\newblock \emph{\bibinfo{journal}{Science}} \textbf{\bibinfo{volume}{378}},
  \bibinfo{pages}{1067--1074} (\bibinfo{year}{2022}).

\bibitem{byrne1994machiavellian}
\bibinfo{author}{Byrne, R.} \& \bibinfo{author}{Whiten, A.}
\newblock \emph{\bibinfo{title}{Machiavellian intelligence}}
  (\bibinfo{publisher}{Oxford Univ. Press}, \bibinfo{year}{1994}).

\bibitem{lanctot2017unified}
\bibinfo{author}{Lanctot, M.} \emph{et~al.}
\newblock \bibinfo{title}{A unified game-theoretic approach to multiagent
  reinforcement learning}.
\newblock \emph{\bibinfo{journal}{Adv. {NeurIPS}}}
  \textbf{\bibinfo{volume}{30}} (\bibinfo{year}{2017}).

\bibitem{vinyals2019grandmaster}
\bibinfo{author}{Vinyals, O.} \emph{et~al.}
\newblock \bibinfo{title}{Grandmaster level in starcraft ii using multi-agent
  reinforcement learning}.
\newblock \emph{\bibinfo{journal}{Nature}} \textbf{\bibinfo{volume}{575}},
  \bibinfo{pages}{350--354} (\bibinfo{year}{2019}).

\bibitem{rendell2010copy}
\bibinfo{author}{Rendell, L.} \emph{et~al.}
\newblock \bibinfo{title}{Why copy others? insights from the social learning
  strategies tournament}.
\newblock \emph{\bibinfo{journal}{Science}} \textbf{\bibinfo{volume}{328}},
  \bibinfo{pages}{208--213} (\bibinfo{year}{2010}).

\bibitem{fang2010balancing}
\bibinfo{author}{Fang, C.}, \bibinfo{author}{Lee, J.} \&
  \bibinfo{author}{Schilling, M.A.}
\newblock \bibinfo{title}{Balancing exploration and exploitation through
  structural design: The isolation of subgroups and organizational learning}.
\newblock \emph{\bibinfo{journal}{Org. Sci.}} \textbf{\bibinfo{volume}{21}},
  \bibinfo{pages}{625--642} (\bibinfo{year}{2010}).

\bibitem{lazer2007network}
\bibinfo{author}{Lazer, D.} \& \bibinfo{author}{Friedman, A.}
\newblock \bibinfo{title}{The network structure of exploration and
  exploitation}.
\newblock \emph{\bibinfo{journal}{Admin. Sci. Quart.}}
  \textbf{\bibinfo{volume}{52}}, \bibinfo{pages}{667--694}
  (\bibinfo{year}{2007}).

\bibitem{mason2008propagation}
\bibinfo{author}{Mason, W.A.}, \bibinfo{author}{Jones, A.} \&
  \bibinfo{author}{Goldstone, R.L.}
\newblock \bibinfo{title}{Propagation of innovations in networked groups.}
\newblock \emph{\bibinfo{journal}{J. Exp. Psychol. Gen.}}
  \textbf{\bibinfo{volume}{137}}, \bibinfo{pages}{422} (\bibinfo{year}{2008}).

\bibitem{vlasceanu2021network}
\bibinfo{author}{Vlasceanu, M.}, \bibinfo{author}{Morais, M.J.} \&
  \bibinfo{author}{Coman, A.}
\newblock \bibinfo{title}{Network structure impacts the synchronization of
  collective beliefs}.
\newblock \emph{\bibinfo{journal}{J. Cog. Cult.}}
  \textbf{\bibinfo{volume}{21}}, \bibinfo{pages}{431--448}
  (\bibinfo{year}{2021}).

\bibitem{coman2016mnemonic}
\bibinfo{author}{Coman, A.}, \bibinfo{author}{Momennejad, I.},
  \bibinfo{author}{Drach, R.D.} \& \bibinfo{author}{Geana, A.}
\newblock \bibinfo{title}{Mnemonic convergence in social networks: The emergent
  properties of cognition at a collective level}.
\newblock \emph{\bibinfo{journal}{Proc. Natl. Acad. Sci.}}
  \textbf{\bibinfo{volume}{113}}, \bibinfo{pages}{8171--8176}
  (\bibinfo{year}{2016}).

\bibitem{centola2022network}
\bibinfo{author}{Centola, D.}
\newblock \bibinfo{title}{The network science of collective intelligence}.
\newblock \emph{\bibinfo{journal}{Trends Cog. Sci.}}  (\bibinfo{year}{2022}).

\bibitem{bernstein2018intermittent}
\bibinfo{author}{Bernstein, E.}, \bibinfo{author}{Shore, J.} \&
  \bibinfo{author}{Lazer, D.}
\newblock \bibinfo{title}{How intermittent breaks in interaction improve
  collective intelligence}.
\newblock \emph{\bibinfo{journal}{Proc. Natl. Acad. Sci.}}
  \textbf{\bibinfo{volume}{115}}, \bibinfo{pages}{8734--8739}
  (\bibinfo{year}{2018}).

\bibitem{mckee2023scaffolding}
\bibinfo{author}{McKee, K.R.} \emph{et~al.}
\newblock \bibinfo{title}{Scaffolding cooperation in human groups with deep
  reinforcement learning}.
\newblock \emph{\bibinfo{journal}{Nat. Hum. Behav.}} \bibinfo{pages}{1--10}
  (\bibinfo{year}{2023}).

\bibitem{osa2018algorithmic}
\bibinfo{author}{Osa, T.} \emph{et~al.}
\newblock \bibinfo{title}{An algorithmic perspective on imitation learning}.
\newblock \emph{\bibinfo{journal}{Found. Trends Robot.}}
  \textbf{\bibinfo{volume}{7}}, \bibinfo{pages}{1--179} (\bibinfo{year}{2018}).

\bibitem{torabi2018behavioral}
\bibinfo{author}{Torabi, F.}, \bibinfo{author}{Warnell, G.} \&
  \bibinfo{author}{Stone, P.}
\newblock \bibinfo{title}{Behavioral cloning from observation}.
\newblock \emph{\bibinfo{journal}{Proc. {IJCAI}}} \bibinfo{pages}{4950--4957}
  (\bibinfo{year}{2018}).

\bibitem{ho2016generative}
\bibinfo{author}{Ho, J.} \& \bibinfo{author}{Ermon, S.}
\newblock \bibinfo{title}{Generative adversarial imitation learning}.
\newblock \emph{\bibinfo{journal}{Adv. {NeurIPS}}}
  \textbf{\bibinfo{volume}{29}} (\bibinfo{year}{2016}).

\bibitem{liu2021motor}
\bibinfo{author}{Liu, S.} \emph{et~al.}
\newblock \bibinfo{title}{From motor control to team play in simulated humanoid
  football}.
\newblock {\bibinfo{journal}{Preprint at https://arXiv.org/abs/2105.12196}}
  (\bibinfo{year}{2021}).

\bibitem{borsa2019observational}
\bibinfo{author}{Borsa, D.} \emph{et~al.}
\newblock \bibinfo{title}{Observational learning by reinforcement learning}.
\newblock \emph{\bibinfo{journal}{Proc. {AAMAS}}} \bibinfo{pages}{1117--1124}
  (\bibinfo{year}{2019}).

\bibitem{ndousse2021emergent}
\bibinfo{author}{Ndousse, K.K.}, \bibinfo{author}{Eck, D.},
  \bibinfo{author}{Levine, S.} \& \bibinfo{author}{Jaques, N.}
\newblock \bibinfo{title}{Emergent social learning via multi-agent
  reinforcement learning}.
\newblock \emph{\bibinfo{journal}{Proc. {ICML}}} \bibinfo{pages}{7991--8004}
  (\bibinfo{year}{2021}).

\bibitem{nisioti2022social}
\bibinfo{author}{Nisioti, E.}, \bibinfo{author}{Mahaut, M.},
  \bibinfo{author}{Oudeyer, P.-Y.}, \bibinfo{author}{Momennejad, I.} \&
  \bibinfo{author}{Moulin-Frier, C.}
\newblock \bibinfo{title}{Social network structure shapes innovation:
  Experience-sharing in RL with SAPIENS}.
\newblock {\bibinfo{journal}{Preprint at https://arXiv.org/abs/2206.05060}}
  (\bibinfo{year}{2022}).

\bibitem{jablonka2006evolution}
\bibinfo{author}{Jablonka, E.} \& \bibinfo{author}{Lamb, M.J.}
\newblock \bibinfo{title}{The evolution of information in the major
  transitions}.
\newblock \emph{\bibinfo{journal}{J. Theor. Biol.}}
  \textbf{\bibinfo{volume}{239}}, \bibinfo{pages}{236--246}
  (\bibinfo{year}{2006}).

\bibitem{henrich2016secret}
\bibinfo{author}{Henrich, J.}
\newblock \emph{\bibinfo{title}{The secret of our success: How culture is
  driving human evolution, domesticating our species, and making us smarter}}
  (\bibinfo{publisher}{Princeton Univ. Press}, \bibinfo{year}{2016}).

\bibitem{bowling2019cell}
\bibinfo{author}{Bowling, S.}, \bibinfo{author}{Lawlor, K.} \&
  \bibinfo{author}{Rodr{\'\i}guez, T.A.}
\newblock \bibinfo{title}{Cell competition: the winners and losers of fitness
  selection}.
\newblock \emph{\bibinfo{journal}{Development}} \textbf{\bibinfo{volume}{146}},
  \bibinfo{pages}{dev167486} (\bibinfo{year}{2019}).

\bibitem{raff1992social}
\bibinfo{author}{Raff, M.C.}
\newblock \bibinfo{title}{Social controls on cell survival and cell death}.
\newblock \emph{\bibinfo{journal}{Nature}} \textbf{\bibinfo{volume}{356}},
  \bibinfo{pages}{397--400} (\bibinfo{year}{1992}).

\bibitem{ferrante2015evolution}
\bibinfo{author}{Ferrante, E.}, \bibinfo{author}{Turgut, A.E.},
  \bibinfo{author}{Du{\'e}{\~n}ez-Guzm{\'a}n, E.}, \bibinfo{author}{Dorigo, M.}
  \& \bibinfo{author}{Wenseleers, T.}
\newblock \bibinfo{title}{Evolution of self-organized task specialization in
  robot swarms}.
\newblock \emph{\bibinfo{journal}{PLoS Comp. Biol.}}
  \textbf{\bibinfo{volume}{11}}, \bibinfo{pages}{e1004273}
  (\bibinfo{year}{2015}).

\bibitem{peysakhovich2018prosocial}
\bibinfo{author}{Peysakhovich, A.} \& \bibinfo{author}{Lerer, A.}
\newblock \bibinfo{title}{Prosocial learning agents solve generalized stag
  hunts better than selfish ones}.
\newblock \emph{\bibinfo{journal}{Proc. {AAMAS}}} \bibinfo{pages}{2043--2044}
  (\bibinfo{year}{2018}).

\bibitem{brambilla2013swarm}
\bibinfo{author}{Brambilla, M.}, \bibinfo{author}{Ferrante, E.},
  \bibinfo{author}{Birattari, M.} \& \bibinfo{author}{Dorigo, M.}
\newblock \bibinfo{title}{Swarm robotics: a review from the swarm engineering
  perspective}.
\newblock \emph{\bibinfo{journal}{Swarm Intell.}} \textbf{\bibinfo{volume}{7}},
  \bibinfo{pages}{1--41} (\bibinfo{year}{2013}).

\bibitem{oroojlooy2022review}
\bibinfo{author}{Oroojlooy, A.} \& \bibinfo{author}{Hajinezhad, D.}
\newblock \bibinfo{title}{A review of cooperative multi-agent deep
  reinforcement learning}.
\newblock \emph{\bibinfo{journal}{Appl. Intell.}} \bibinfo{pages}{1--46}
  (\bibinfo{year}{2022}).

\bibitem{schranz2020swarm}
\bibinfo{author}{Schranz, M.}, \bibinfo{author}{Umlauft, M.},
  \bibinfo{author}{Sende, M.} \& \bibinfo{author}{Elmenreich, W.}
\newblock \bibinfo{title}{Swarm robotic behaviors and current applications}.
\newblock \emph{\bibinfo{journal}{Front. Robot. {AI}}}
  \textbf{\bibinfo{volume}{7}}, \bibinfo{pages}{36} (\bibinfo{year}{2020}).

\bibitem{leibo2021scalable}
\bibinfo{author}{Leibo, J.Z.} \emph{et~al.}
\newblock \bibinfo{title}{Scalable evaluation of multi-agent reinforcement
  learning with {M}elting {P}ot}.
\newblock \emph{\bibinfo{journal}{Proc. {ICML}}} \bibinfo{pages}{6187--6199}
  (\bibinfo{year}{2021}).

\bibitem{sunehag2023diversity}
\bibinfo{author}{Sunehag, P.}, \bibinfo{author}{Vezhnevets, A.S.},
  \bibinfo{author}{Du{\'e}{\~n}ez-Guzm{\'a}n, E.}, \bibinfo{author}{Mordach,
  I.} \& \bibinfo{author}{Leibo, J.Z.}
\newblock \bibinfo{title}{Diversity through exclusion ({DTE}): Niche
  identification for reinforcement learning through value-decomposition}.
\newblock \emph{\bibinfo{journal}{Proc. {AAMAS}}} \bibinfo{pages}{2827--2829}
  (\bibinfo{year}{2023}).

\bibitem{wang2019evolving}
\bibinfo{author}{Wang, J.X.} \emph{et~al.}
\newblock \bibinfo{title}{Evolving intrinsic motivations for altruistic
  behavior}.
\newblock \emph{\bibinfo{journal}{Proc. {AAMAS}}} \bibinfo{pages}{683--692}
  (\bibinfo{year}{2019}).

\bibitem{gemp2022d3c}
\bibinfo{author}{Gemp, I.} \emph{et~al.}
\newblock \bibinfo{title}{{D3C}: Reducing the price of anarchy in multi-agent
  learning}.
\newblock \emph{\bibinfo{journal}{Proc. {AAMAS}}} \bibinfo{pages}{498--506}
  (\bibinfo{year}{2022}).

\bibitem{zheng2022ai}
\bibinfo{author}{Zheng, S.}, \bibinfo{author}{Trott, A.},
  \bibinfo{author}{Srinivasa, S.}, \bibinfo{author}{Parkes, D.C.} \&
  \bibinfo{author}{Socher, R.}
\newblock \bibinfo{title}{The {AI} economist: Taxation policy design via
  two-level deep multiagent reinforcement learning}.
\newblock \emph{\bibinfo{journal}{Sci. Adv.}} \textbf{\bibinfo{volume}{8}},
  \bibinfo{pages}{eabk2607} (\bibinfo{year}{2022}).

\bibitem{koster2022human}
\bibinfo{author}{Koster, R.} \emph{et~al.}
\newblock \bibinfo{title}{Human-centered mechanism design with democratic
  {AI}}.
\newblock \emph{\bibinfo{journal}{Nat. Hum. Behav.}} \textbf{\bibinfo{volume}{6}},
  \bibinfo{pages}{1398--1407} (\bibinfo{year}{2022}).

\bibitem{dean2012identification}
\bibinfo{author}{Dean, L.G.}, \bibinfo{author}{Kendal, R.L.},
  \bibinfo{author}{Schapiro, S.J.}, \bibinfo{author}{Thierry, B.} \&
  \bibinfo{author}{Laland, K.N.}
\newblock \bibinfo{title}{Identification of the social and cognitive processes
  underlying human cumulative culture}.
\newblock \emph{\bibinfo{journal}{Science}} \textbf{\bibinfo{volume}{335}},
  \bibinfo{pages}{1114--1118} (\bibinfo{year}{2012}).

\bibitem{muthukrishna2016innovation}
\bibinfo{author}{Muthukrishna, M.} \& \bibinfo{author}{Henrich, J.}
\newblock \bibinfo{title}{Innovation in the collective brain}.
\newblock \emph{\bibinfo{journal}{Phil. Trans. Royal Soc. B}}
  \textbf{\bibinfo{volume}{371}}, \bibinfo{pages}{20150192}
  (\bibinfo{year}{2016}).

\bibitem{dunbar2017there}
\bibinfo{author}{Dunbar, R.I.} \& \bibinfo{author}{Shultz, S.}
\newblock \bibinfo{title}{Why are there so many explanations for primate brain
  evolution?}
\newblock \emph{\bibinfo{journal}{Philos. Trans. Royal Soc. B}}
  \textbf{\bibinfo{volume}{372}}, \bibinfo{pages}{20160244}
  (\bibinfo{year}{2017}).

\bibitem{kirby2015compression}
\bibinfo{author}{Kirby, S.}, \bibinfo{author}{Tamariz, M.},
  \bibinfo{author}{Cornish, H.} \& \bibinfo{author}{Smith, K.}
\newblock \bibinfo{title}{Compression and communication in the cultural
  evolution of linguistic structure}.
\newblock \emph{\bibinfo{journal}{Cognition}} \textbf{\bibinfo{volume}{141}},
  \bibinfo{pages}{87--102} (\bibinfo{year}{2015}).

\bibitem{ostrom2005understanding}
\bibinfo{author}{Ostrom, E.}
\newblock \emph{\bibinfo{title}{Understanding institutional diversity}}
  (\bibinfo{publisher}{Princeton Univ. Press},
  \bibinfo{year}{2005}).

\bibitem{havrylov2017emergence}
\bibinfo{author}{Havrylov, S.} \& \bibinfo{author}{Titov, I.}
\newblock \bibinfo{title}{Emergence of language with multi-agent games:
  Learning to communicate with sequences of symbols}.
\newblock \emph{\bibinfo{journal}{Adv. {NeurIPS}}}
  \textbf{\bibinfo{volume}{30}} (\bibinfo{year}{2017}).

\bibitem{mordatch2018emergence}
\bibinfo{author}{Mordatch, I.} \& \bibinfo{author}{Abbeel, P.}
\newblock \bibinfo{title}{Emergence of grounded compositional language in
  multi-agent populations}.
\newblock \emph{\bibinfo{journal}{Proc. {AAAI} Conf. Artif. Intell.}}
  \textbf{\bibinfo{volume}{32}} (\bibinfo{year}{2018}).

\bibitem{brown2020language}
\bibinfo{author}{Brown, T.} \emph{et~al.}
\newblock \bibinfo{title}{Language models are few-shot learners}.
\newblock \emph{\bibinfo{journal}{Adv. {NeurIPS}}}
  \textbf{\bibinfo{volume}{33}}, \bibinfo{pages}{1877--1901}
  (\bibinfo{year}{2020}).

\bibitem{chowdhery2022palm}
\bibinfo{author}{Chowdhery, A.} \emph{et~al.}
\newblock \bibinfo{title}{PaLM: Scaling language modeling with pathways}.
\newblock {\bibinfo{journal}{Preprint at https://arXiv.org/abs/2204.02311}}
  (\bibinfo{year}{2022}).

\bibitem{chan2022data}
\bibinfo{author}{Chan, S.C.} \emph{et~al.}
\newblock \bibinfo{title}{Data distributional properties drive emergent
  few-shot learning in transformers}.
\newblock \emph{\bibinfo{journal}{Adv. {NeurIPS}}}
  \textbf{\bibinfo{volume}{35}}, \bibinfo{pages}{18878--18891}
  (\bibinfo{year}{2022}).

\bibitem{wei2022chain}
\bibinfo{author}{Wei, J.} \emph{et~al.}
\newblock \bibinfo{title}{Chain of thought prompting elicits reasoning in large
  language models}.
\newblock \emph{\bibinfo{journal}{Adv. {NeurIPS}}}
  \textbf{\bibinfo{volume}{35}}, \bibinfo{pages}{24824--24837}
  (\bibinfo{year}{2022}).

\bibitem{bisk2020experience}
\bibinfo{author}{Bisk, Y.} \emph{et~al.}
\newblock \bibinfo{title}{Experience grounds language}.
\newblock \emph{\bibinfo{journal}{Proc. {EMNLP}}} \bibinfo{pages}{8718--8735}
  (\bibinfo{year}{2020}).

\bibitem{ullman2023large}
\bibinfo{author}{Ullman, T.}
\newblock \bibinfo{title}{Large language models fail on trivial alterations to
  theory-of-mind tasks}.
\newblock {\bibinfo{journal}{Preprint at https://arXiv.org/abs/2302.08399}}
  (\bibinfo{year}{2023}).

\bibitem{liu2022mind}
\bibinfo{author}{Liu, R.} \emph{et~al.}
\newblock \bibinfo{title}{Mind's eye: Grounded language model reasoning through
  simulation}.
\newblock \emph{\bibinfo{journal}{Proc. {ICLR}}}
  (\bibinfo{year}{2023}).

\bibitem{glaese2022improving}
\bibinfo{author}{Glaese, A.} \emph{et~al.}
\newblock \bibinfo{title}{Improving alignment of dialogue agents via targeted
  human judgements}.
\newblock {\bibinfo{journal}{Preprint at https://arXiv.org/abs/2209.14375}}
  (\bibinfo{year}{2022}).

\bibitem{colas2022language}
\bibinfo{author}{Colas, C.}, \bibinfo{author}{Karch, T.},
  \bibinfo{author}{Moulin-Frier, C.} \& \bibinfo{author}{Oudeyer, P.-Y.}
\newblock \bibinfo{title}{Language and culture internalization for human-like
  autotelic ai}.
\newblock \emph{\bibinfo{journal}{Nat. Mach. Intell.}}
  \textbf{\bibinfo{volume}{4}}, \bibinfo{pages}{1068--1076}
  (\bibinfo{year}{2022}).

\bibitem{villalobos2022will}
\bibinfo{author}{Villalobos, P.} \emph{et~al.}
\newblock \bibinfo{title}{Will we run out of data? an analysis of the limits of
  scaling datasets in machine learning}.
\newblock {\bibinfo{journal}{Preprint at https://arXiv.org/abs/2211.04325}}
  (\bibinfo{year}{2022}).

\bibitem{gazda2016driver}
\bibinfo{author}{Gazda, S.K.}
\newblock \bibinfo{title}{Driver-barrier feeding behavior in bottlenose
  dolphins (tursiops truncatus): New insights from a longitudinal study}.
\newblock \emph{\bibinfo{journal}{Marine Mammal Sci.}}
  \textbf{\bibinfo{volume}{32}}, \bibinfo{pages}{1152--1160}
  (\bibinfo{year}{2016}).

\bibitem{bales2021pair}
\bibinfo{author}{Bales, K.L.} \emph{et~al.}
\newblock \bibinfo{title}{What is a pair bond?}
\newblock \emph{\bibinfo{journal}{Horm. Behav.}}
  \textbf{\bibinfo{volume}{136}}, \bibinfo{pages}{105062}
  (\bibinfo{year}{2021}).

\bibitem{lukas2018social}
\bibinfo{author}{Lukas, D.} \& \bibinfo{author}{Clutton-Brock, T.}
\newblock \bibinfo{title}{Social complexity and kinship in animal societies}.
\newblock \emph{\bibinfo{journal}{Ecol. Lett.}} \textbf{\bibinfo{volume}{21}},
  \bibinfo{pages}{1129--1134} (\bibinfo{year}{2018}).

\bibitem{feldman2015adaptive}
\bibinfo{author}{Feldman, R.}
\newblock \bibinfo{title}{The adaptive human parental brain: implications for
  children's social development}.
\newblock \emph{\bibinfo{journal}{Trends Neurosci.}}
  \textbf{\bibinfo{volume}{38}}, \bibinfo{pages}{387--399}
  (\bibinfo{year}{2015}).

\bibitem{tarr2015synchrony}
\bibinfo{author}{Tarr, B.}, \bibinfo{author}{Launay, J.},
  \bibinfo{author}{Cohen, E.} \& \bibinfo{author}{Dunbar, R.}
\newblock \bibinfo{title}{Synchrony and exertion during dance independently
  raise pain threshold and encourage social bonding}.
\newblock \emph{\bibinfo{journal}{Biol. Lett.}} \textbf{\bibinfo{volume}{11}},
  \bibinfo{pages}{20150767} (\bibinfo{year}{2015}).

\bibitem{lieberwirth2014social}
\bibinfo{author}{Lieberwirth, C.} \& \bibinfo{author}{Wang, Z.}
\newblock \bibinfo{title}{Social bonding: Regulation by neuropeptides}.
\newblock \emph{\bibinfo{journal}{Front. Neurosci.}}
  \textbf{\bibinfo{volume}{8}}, \bibinfo{pages}{171} (\bibinfo{year}{2014}).

\bibitem{aagren2019enforcement}
\bibinfo{author}{{\AA}gren, J.A.}, \bibinfo{author}{Davies, N.G.} \&
  \bibinfo{author}{Foster, K.R.}
\newblock \bibinfo{title}{Enforcement is central to the evolution of
  cooperation}.
\newblock \emph{\bibinfo{journal}{Nat. Ecol. Evol.}}
  \textbf{\bibinfo{volume}{3}}, \bibinfo{pages}{1018--1029}
  (\bibinfo{year}{2019}).

\bibitem{wilkins2014domestication}
\bibinfo{author}{Wilkins, A.S.}, \bibinfo{author}{Wrangham, R.W.} \&
  \bibinfo{author}{Fitch, W.T.}
\newblock \bibinfo{title}{The ``domestication syndrome'' in mammals: A
  unified explanation based on neural crest cell behavior and genetics}.
\newblock \emph{\bibinfo{journal}{Genetics}} \textbf{\bibinfo{volume}{197}},
  \bibinfo{pages}{795--808} (\bibinfo{year}{2014}).

\end{thebibliography}
\end{document}